\documentclass[graybox]{article}

\usepackage{type1cm}        
\usepackage{makeidx}         
\usepackage{graphicx}        
\usepackage{multicol}        
\usepackage[bottom]{footmisc}

\usepackage{newtxtext}       %
\usepackage{newtxmath}       
\usepackage{todonotes}
\usepackage{url}

\makeindex             

\begin{document}

\title{AutoML @ NeurIPS 2018 challenge: Design and Results\thanks{Preprint submitted to NeurIPS2018 Volume of Springer Series on Challenges in Machine Learning}}
\author{Hugo Jair Escalante$^{1,2}$, Wei-Wei Tu$^{3}$, Isabelle Guyon$^{2,4}$,  Daniel L. Silver$^{5}$,\\ Evelyne Viegas$^{6}$, Yuqiang Chen$^{3}$, Wenyuan Dai$^{3}$, Qiang Yang$^{7}$\\\\
$1$ Instituto Nacional de Astrof\'isica, \'Optica, y Electr\'onica, Mexico,\\ 
$2$ ChaLearn, Berkeley, CA, USA, $3$ 4Paradigm, Beijing, China\\
$4$ UPSud/INRIA Universit\'e Paris-Saclay, France\\
$5$ Acadia University, CA, $6$ Microsoft Research, USA\\
$7$ Hong Kong University of Science and Technology, Hong Kong\\
\tt{hugojair@inaoep.mx}
}
%
%
\maketitle

 \abstract{We organized a  competition on Autonomous Lifelong Machine Learning with Drift that was part of the competition program of NeurIPS 2018. This data driven competition asked participants to develop computer programs capable of solving supervised learning problems where the i.i.d. assumption did not hold. Large data sets were arranged in a lifelong learning and evaluation scenario and CodaLab was used as the challenge platform. The challenge attracted more than 300 participants in its two month duration. This chapter describes the design of the challenge and summarizes its main results. 
 }

\section{Introduction}
\label{sec:intro}
Machine learning has achieved great successes in online advertising, recommender systems, financial market analysis, computer vision, computational linguistics, bioinformatics and many other fields.  However, its success crucially relies on human machine learning experts, as human experts are involved to some extent, in all systems design stages. In fact,  it is still  common for humans to take critical decisions in aspects like:  converting a real world problem into a machine learning one, data gathering, formatting and preprocessing, feature engineering, selecting or designing model architectures, hyper-parameter tuning, assessment of model performance, deploying  on-line ML systems, among others. The complexity of these tasks, which is often beyond non-experts, together with the rapid growth of applications, have motivated a demand for off-the-shelf machine learning methods that can be used easily and without any expert knowledge. 

The field of study dealing the  progressive automation of machine learning is now referred to AutoML (Automatic Machine Learning or Automated Machine Learning)~\cite{AutoMLbookCIML}.  Although the term is somewhat new, this topic has been studied for a while in the context of machine learning and it is even \emph{older} in related fields like portfolio and algorithm design/analysis~\cite{RICE197665}.

Following the success of two previous international challenges in AutoML  and numerous hackathons we co-organized (see \url{http://automl.chalearn.org/}), which attracted hundreds of participants, we organized a competition on the topic of  \emph{Autonomous Lifelong Machine Learning with Drift} that was collocated with the Neural Information and Processing Systems Conference (NeurIPS) 2018. The organized competition, which was abbreviated as {\bf AutoML for Lifelong Machine Learning}, or {\bf AutoLML}, challenged participants to design a computer program capable of providing solutions to supervised learning problems autonomously (without any user intervention).  Compared to previous competitions we organized, our new focus is on {\bf drifting concepts}, getting away from the simpler {\em i.i.d.} cases we were previously confined to, and the scale of tasks with {\bf datasets much larger} than previously made available to participants.

As in previous editions, participants were required to submit code that was autonomously evaluated in the challenge platform. All participants had access to the same computing and storage resources, making the evaluation fair. The challenge was open for about 2 months and attracted more than 300 participants. Interesting findings and open issues were the most important outcome from this competition, both, from the organization and participant perspectivs. Overall, the challenge was a success, and this has motivated the development of future related challenges in the LML area. This chapter provides a comprehensive description of the competition, including the definition of the task scenario, evaluation protocol and an overview of results. 

The remainder of this paper is organized as follows. 
Next section elaborates on the relevance of the challenge and presents the considered setting. Section~\ref{sec:challenge} describes the challenge in detail, introducing the data sets, evaluation metrics and protocol. Next, Section~\ref{sec:results} summarizes the results of the challenge and analyzes the top ranked submissions. Finally, Section~\ref{sec:discussion} outlines conclusions and explains future research directions on AutoML.

\section{Challenge setting and background}
\label{sec:setting}

The autonomous all-problem machine learning method has been a dream for machine learning researchers for a long time and recently there have been advances in the area (see e.g.,~\cite{algosel}). 
However, a lack of suitable benchmarks, evaluation protocols and performance metrics has limited progress. Recently, we have organized challenges on Autonomous Machine Learning that have made significant progress in the field, see e.g.,~\cite{DBLP:conf/icml/GuyonCEEJLMRRSS16,DBLP:conf/ijcnn/2015}. The challenges have attracted a large number of participants (almost 1,000 in the combined challenges), providing evidence of the relevance of the problem and the interest from the community. In this recently organized challenge, we aimed to explore areas of AutoML that have not been studied so far, and that are present in almost every possible application of AutoML.  The novel components of the proposed challenge are: the use of \textbf{large scale} datasets coming from \textbf{real-world applications}, where data are subject to the \textbf{concept drift} phenomenon and from adopting a \textbf{lifelong} learning and evaluation framework. 
In the remainder of this section we briefly review background information, summarize previous related events and highlight the elements of novelty of the AutoML at NeurIPS2018 challenge.

\subsection{Concept drift and lifelong learning}
We consider the AutoML problem in the context of supervised learning, specifically, we consider binary classification problems. Different from previous challenges we have organized, the AutoML at NeurIPS2018 challenge focused on problems with presence of concept drift~\cite{schlimmer:aaai86}.  The problem of concept drift in predictive analytics is related to time series prediction, but at a different scale of granularity: it typically addresses problems in which the data distribution is changing relatively slowly. Batches of data may be arriving daily, weekly, monthly, or yearly, for instance. In many cases, only the order of arrival matters and the exact timing is not recorded. This setting poses the problem of continuously adapting learning machines (Lifelong learning). We tackle multi-variate problems with a large number of features and a single (binary) target. Typical tasks include customer relationship management, on-line advertising, recommendation, sentiment analysis, fraud detection, spam filtering, transportation monitoring, econometrics, patient monitoring, climate monitoring, and manufacturing. 
 For the organized challenge we made available new large scale datasets associated with such real-world applications, see Section~\ref{sec:data}. 

To the best of our knowledge, AutoLML is the first Lifelong Machine Learning challenge to be organized. And its timing was good, because LML is increasingly attractive to the machine learning community. Particularly, since DARPA announced a related program called Lifelong Learning Machines \footnote{\url{https://www.darpa.mil/news-events/2017-03-16}} many teams around the world have started work on the topic with applications ranging from data mining to robotics. Our recently organized challenge focused on binary concept learning as a starting point for such LML competitions. 
The proposed challenge is relatively conservative, in the continuity of previous challenges, not addressing (yet) decision processes and reinforcement learning. However, our setting has great practical importance to the industry.
%

Autonomous Lifelong ML differs from traditional AutoML in the sense that the learning machines keep acquiring knowledge from every task they are exposed to in order to beneficially bias the development better models for new  tasks~\cite{danny13,DBLP:series/synthesis/2016Chen}. This is very much related to ``transfer learning''\cite{Pan-2010}, the problem of generalizing from task to task. The difference lies in the long term and continuous aspect of Lifelong ML~\cite{danny13}. 

\subsection{Previously related competitions}
One of the oldest challenges organized on a related topic is the Pascal 2 EU network of excellence challenge on ``covariate shift'', organized in 2005~\cite{candela-2005}. The challenge is no longer on-line. It featured training sets and test sets differently distributed. Organizing the AutoML at NeurIPS2018 challenge on a related theme more than ten years later  pushed the community to update the state-of-the-art. 

The authors  were also involved in the organization of an ``unsupervised and transfer learning'' challenge (ICML and IJCNN 2011)~\cite{guyon-2011}. In that setting the participants were exposed to tasks drawn from the same datasets, but with different subsets of the labels ({\em e.g. for image classification, one task was animal classification, the other vehicle classification}). However, this competition did not involve  concept drift.

We have organized two AutoML challenges in the past. The first one lasted 2 years and comprised 6 stages of increasing difficulty, with a total of 30 datasets. A variety of supervised learning tasks were considered, and we evaluated the benefit of purely AutoML solutions {\em vs.} standard parameter tuning techniques. The findings of the challenge were presented in NIPS2015\footnote{\url{http://ciml.chalearn.org/home/schedule}}, IJCNN2015~\cite{DBLP:conf/ijcnn/2015}, IJCNN2016\footnote{\url{http://www.wcci2016.org/programs.php?id=home}} and ICML2016~\cite{DBLP:conf/icml/GuyonCEEJLMRRSS16}. More than 600 participants registered for the competition and interesting findings were drawn from this challenge, see~\cite{DBLP:conf/ijcnn/2015,DBLP:conf/icml/GuyonCEEJLMRRSS16}.  More recently, we organized an AutoML challenge that was part of the PAKDD2018 competition program\footnote{\url{https://www.4paradigm.com/competition/pakdd2018}}. In this challenge larger datasets coming from real-world problems, including additional feature types were considered. The challenge lasted fro three months and attracted almost 300 participants. 
The increasing participation, interesting findings and discovery of related problems motivated us to organize this new AutoLML challenge. 

\subsection{Elements of novelty of this challenge}
The success of our previous AutoML challenges can be found at ~\cite{DBLP:conf/ijcnn/2015,DBLP:conf/icml/GuyonCEEJLMRRSS16,automlchallenges} and the respective websites are still publicly available for review and comparison.  With this new AutoLML challenge we aimed to move to large, real-world data sets where lifelong learning and evaluation over time was necessary. Although the models focused on binary classification problems, the  challenge introduced the following difficulties that made it unique and relevant for the machine learning community:
\begin{enumerate}
\item  \textbf{Concept drift.} We address binary classification problems (not time series prediction), 
but all the instances in the datasets were {\bf provided in chronological order}. In this way, slow changes in data distribution could be exploited.
%
%
	\item \textbf{Lifelong Evaluation}. We got away from the classical data split into a training set and a test set.
Data were cut into blocks respecting the time ordering, and fed to the AutoLML code submitted by the participants block by block (see Figure~\ref{fig:scenario}). Starting with an initial block of labeled data for training, we revealed the labels of subsequent blocks only after predictions have been made on them (evaluation on a sliding window). The final score was the average performance over all blocks.
	\item \textbf{Targeted real-world tasks.} In  previous competitions we invited  participants to build AutoML systems for a wide range of application domains. Although this is the right direction for AutoML, we felt it would be too difficult a start for LML. Therefore, in this initial LML competition, we focused on tasks relatively similar to each other, taken from business decision applications, such as on-line advertising, recommendation systems.
	\item \textbf{Large scale data sets.} In previous competitions, we only offered tens of thousands of instances for each dataset. In this competition, we  used datasets with sizes ranging from hundreds of thousands to millions of instances making it closer to real-world scenarios. This  enforced participants to care more about the efficiency of their AutoML strategies due to the limited time budget they had for each dataset. 
	\item \textbf{Diverse features.} There were more attribute types in this competition, compared to previous ones where we only had single valued numerical/categorical attributes. This time multi-valued and temporal attributes will be considered. These attribute types are commonly seen in real-world scenarios, where different preprocessing methods lead to different results. This emphasizes that automatic feature engineering is still important in real-world scenarios.
    \item \textbf{Tough categorical features.} In these real-world applications, categorical features with a large number of distinct values are commonly seen. Their frequencies are always following a power-law distribution. These features ({\em e.g., userId, itemId, etc.}) have proven to be useful in these applications. But they are not easy to use for ML beginners. If such features are not encoded properly (e.g., via one hot encoding) 
    the input for ML system will be very high dimensional and sparse, this is quite challenging for AutoML systems.	
%
%
\end{enumerate}

\subsection{Discussion}
Compared to previous competitions on AutoML, the AutoLML challenge focuses on a novel setting with practical implications. A lifelong learning and evaluation perspective made this challenge relevant to research agencies such as DARPA, companies/products such as Automated ML\footnote{\url{https://docs.microsoft.com/en-us/azure/machine-learning/service/concept-automated-ml}} and Cloud AutoML\footnote{\url{https://cloud.google.com/automl/}}, and academic research~\cite{jin2018efficient,autosklearn,AutoMLbookCIML}.  Subsequently, the challenge attracted the attention from the machine learning and data mining communities and resulted in novel and highly effective solutions. Hence, the organized challenge was a success that will motivate further research in the forthcoming years. 

\section{Competition description}
\label{sec:challenge}
\begin{figure}[htb]
\begin{center}
\includegraphics[width=0.95\linewidth]{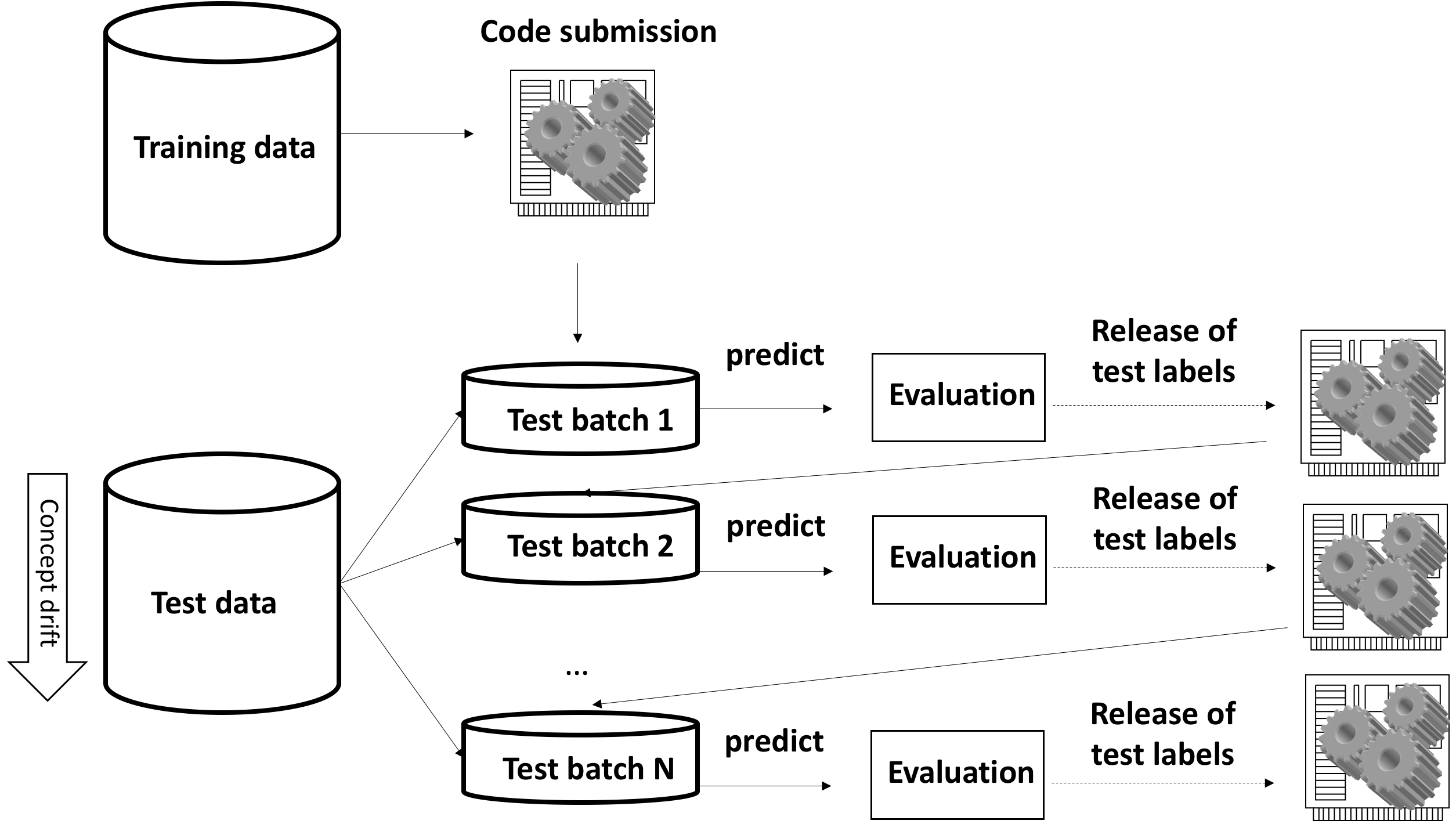}
\caption{Proposed evaluation scenario.}
\label{fig:scenario}
\end{center}
\end{figure}

We adopted an evaluation framework that aimed at assessing the robustness of methods to concept drift and its lifelong learning capabilities. Participants were provided with a set of public datasets (labeled training data and unlabeled test data), and they had to develop their AutoLML solutions. Provided datasets are temporally dependent and are subject to a form of the concept drift phenomenon. The datasets were also split into sequential blocks of data that was first seen as test data and then, with revealed target values, became additonal training data.  See Section~\ref{section:protocol} and Figure~\ref{fig:scenario} for details.

The AutoLML challenge was divided into two stages. In the first "Feedback" phase, participants were able to upload code to the challenge platform, and this code was evaluated on unseen data.  Then participants received immediate feedback on the performance of their code in public datasets. For the second "Final" phase, the last submitted code per participant was be autonomously evaluated on fresh new data that was  kept private for the duration of the competition.  
The remainder of this section provides details on the datasets used and evaluation protocol of the challenge. 

\subsection{Data}
\label{sec:data}
For the challenge we considered 10 datasets that were used in the two phases described in the prior section. Half of the datasets were  made  available to participants (labeled development data and unlabeled test data) during the feedback phase, and the other half remained private and it was used for the evaluation in the Final phase. 
Participants could  develop their AutoLML systems by using the public datasets. During the challenge, participants were able to submit their code and receive feedback on the performance of their solution on unlabeled data released in the feedback phase. 
As for the private datasets, 
participants could not access these datasets, not even the training partitions. The final score was  evaluated on the private datasets. 

The sizes of the datasets differ from hundreds of thousands to millions, this is  10-100 times larger than the previous ones. Table~\ref{tab:datasets} shows some characteristics of the datasets that were considered for feedback phase of the challenge. Datasets used in the private final phase were similarly distributed as those illustrated in the table. Please note that since the private data is still being used in a rematch phase of this challenge\footnote{\url{https://www.4paradigm.com/competition/pakdd2019}} we cannot give details on the features of such data sets.  
\begin{table}[htb]
	\centering
	\caption{Datasets considered for the feedback phase of the AutoML challenge}
	\label{tab:datasets}
	\begin{tabular}{c|c|c|c|c|c|c|c}
		\hline
		Dataset&Budget (s)&\multicolumn{4}{|c|}{No. of Features per type} & No. Features& No. Instances\\
		&  & Cat &	Num	&  MVC	&  Time	& &\\\hline
        A&	3,600&	51&	23&	6&	2&	82&	$\approx$ 10M\\
        B&	600&	17&	7&	1&	0&	25&	$\approx$ 1.9M\\
        C&	1,200&	44&	20&	9&	6&	79&	$\approx$ 2M\\
        D&	600&	17&	54&	1&	4&	76&	$\approx$ 1.5M\\
        E&	1,800&	25&	6&	1&	2&	34&	$\approx$ 17M\\
	\end{tabular}
\end{table}

From Table~\ref{tab:datasets} it can be inferred the complexity of the associated classification tasks. All of the datasets included categorical variables, in fact categorical and multi-valued categorical are majority features for all data sets. The total number of features is reasonable, however the number of samples surpass by a large margin the sizes of datasets considered for previous challenges. 

Data were provided to participants in two different ways. We provided all of the data in tabular format so that participants can focus on the classifier design process. In this  way, time features were converted to integers, whereas categorical and multi-valued categorical variables were encoded using the ordinal encoding implementation from a public library\footnote{\url{http://contrib.scikit-learn.org/categorical-encoding/}}; where our choice for this encoding method was because was the fastest encoding method in the considered library. All the features were concatenated, forming large matrices of samples vs. features.   Since this  representation of the data might be too restrictive for participants, we also provided all features in raw format. In this way, participants could  design their own feature extraction methods. For the distribution of features in different formats we launched two competition sites within the CodaLab platform (see Section~\ref{sec:platform} for more details). 
Each of the datasets was split into blocks respecting the time ordering for the Lifelong evaluation associated to the challenge (see Section~\ref{section:protocol} for details). 




\subsection{Platform and budget}
\label{sec:platform}
The challenge was run on the CodaLab platform\footnote{\url{http://competitions.codalab.org}}.  CodaLab  is an open source platform for hosting competitions. The platform has been jointly developed by Microsoft research, Stanford University and ChaLearn. CodaLab has been succeesfully used  for the AutoML 2015-2018 challenge series. 

The organization of this competition was challenging in several ways, this was by far the competition with largest datasets ever launched in codalab, which make us to find several bugs in the platform. Debugging and improvement of the platform was a collateral result of the organization of this challenge.

In addition to presenting the characteristic of the datasets, Table~\ref{tab:datasets} shows the assigned budget for each of the task. This budget specifies the time in seconds that an AutoML solution can spend in processing each dataset. Failure to adhere to this budget would penalize the submission. The total time given to an AutoML solution to complete the task was a bit more than 2 hours, with the largest data set (A) being allowed one hour to be processed.  The way this budget was determined was by estimating the time taken by the baselines provided to participants in the starting kit. The provided baseline took between one-third and one-quarter the time shown in the budget. Please note that proportional time budgets we associated to data in the final phase.  Each participant was able to upload 2 submissions per day, this mechanism was adopted to avoid saturation of the workers and overfitting  of the feedback phase. 

Each submission was assigned a virtual machine with identical characteristics: 8CPU with 16GB RAM and 100GB of storage capacity. This guaranteed a fair comparison of participants solutions. A queue of six compute workers with these specifications was used for the challenge. Compute workers were sponsored by Univesit\'e Paris Saclay and Microsoft Research throughout a Azure grant. 

For this competition two different environments were provided to participants. Initially a docker container with Python 2.7 and standard machine learning libraries was provided. For this configuration data was provided in tabular form and participants could focus on building a predictive model from the beginning. We refer to this competition setting as Python2\footnote{\url{https://competitions.codalab.org/competitions/19836}}. After receiving feedback from participants we decided to launch a second docker container, this time running Python 3 and allowing the installation of more recent packages and libraries. In addition, we did several improvements to the Python2 release, including: providing raw-features, providing more accurate spend-time and feature-type information to participants solutions. We called this competition setting as Python-3\footnote{\url{https://competitions.codalab.org/competitions/20203}}. Although both environments used different versions of programming language and data formatting, they were using exactly the same challenge settings, including, datasets, evaluation metrics, time budget, etc. Participants were asked to participate in only one of the two settings to avoid people getting advantage of uploading more submissions (otherwise they could submit 2 code entries to each environment).

\subsection{Evaluation metrics}
%
%

As previously mentioned, each dataset was  split into blocks and the data were progressively presented to the participants' classifiers as follows:\\\\
{\footnotesize
\begin{tabular}{| l | l| l | l | }
  \hline	
  STEP \# &	TRAINING DATA  & TEST DATA \\
  \hline
  1 & LABELED BLOCK\_0 &  UNLABELED BLOCK\_1 \\
  2 & LABELED (BLOCK\_0 +  BLOCK\_1) & UNLABELED BLOCK\_2 \\
  3 & LABELED (BLOCK\_0 +  BLOCK\_1 +BLOCK\_2) & UNLABELED BLOCK\_3 \\
  $\cdots$ & $\cdots$  & $\cdots$  \\
  N & EVERYTHING LABELED UP TO BLOCK\_(N-1) & UNLABELED BLOCK\_N \\
  \hline  
\end{tabular}
  }

\normalsize{
\begin{itemize}
\item For each block of each dataset, the evaluation consisted in computing the area under the ROC curve (AUC).
\item For each dataset, we  averaged  the AUCs over all the blocks of the dataset. A ranking was generated  according to this metric.
\item For the final score, we used the average rank over all  datasets.
\item There was a time budget associated to each dataset. Code exceeding the maximum execution time was aborted and the corresponding submission was disqualified. 
\end{itemize}
}
In case of ties, duration of the solution was considered as a secondary criterion. 

\subsection{Baselines and code available}

 We provided a starting kit 
 to participants with the baseline method included. The starting kit allowed participants to make experiments offline following the considered lifelong  evaluation protocol. For offline experimentation, participants were provided with public datasets, not considered in the feedback of final phases.   
 
The baseline provided to participants was an ensemble model based on the gradient boosting implementation from scikit-learn toolkit~\cite{scikitlearn}. The ensemble was incrementally updated (adding more weak learners) as more labeled instances were available to it. At the beginning the classifier was trained with the all labeled data using $l-$weak learners, after each batch of data was released another $h-$weak learners trained on the current batch were added. Since the datasets are very large and the budget was limited, we performed a subsampling of the available labeled patterns in each stage of the training process. Although the performance of this simple baseline was low, it was very useful as a template for participants to develop their own models. In fact, solutions form participants are similar in spirit to the released baseline. Please note that another more effective baseline was also developed by the team in~\cite{madrid:hal-01966962}, however this method was too much computationally demanding as to run under the available computing resources. Participants  had to outperform the baseline in order to claim prizes. 
%
%





\subsection{Evaluation protocol} \label{section:protocol}
The overall adopted evaluation scheme is depicted in Figure~\ref{fig:scenario}. Recall this is a competition receiving \textbf{code submissions}. Participants must prepare an AutoML program to be uploaded to the challenge platform. The code was  executed in computer workers, autonomously;  and allowed to run for a maximum amount of time. Code exceeding this time was  penalized with setting the dataset's AUC as 0, and the overall submission was be disqualified. Different from previous challenges, in this competition we  evaluated the Lifelong learning capabilities of AutoML solutions, hence an appropriate protocol has been designed. 
%
%

The datasets were split into different blocks, each block  representing a stage of the lifelong evaluation scenario. After submission, code submitted by participants  used training data to generate a model, which  then was used to predict labels for the first test set. The performance on this test set was recorded. After this, the labels of the first test set were made available to the submitted code. The code may use such labels to improve its initial model and make predictions for the subsequent test set. The lifelong learning process  continued until all of the test sets were  evaluated. 
%
%

As previously mentioned the challenge comprises two phases:
\begin{itemize}
\item \textbf{Feedback phase.} Public data sets were provided in this phase, where labeled training data and unlabeled test data were released. During this phase, participants could submit code to the challenge platform and they  received immediate feedback (through a leaderboard) on the lifelong performance of their methods on the test set (according to the framework depicted in Figure~\ref{fig:scenario}). Participants could improve their code submissions according to their performance in the leaderboard. The number of submissions that participants could send per day was limited to two.  

\item \textbf{Final phase.} The last code submission from the feedback was automatically migrated to the final phase. Code was then evaluated on 5 new datasets that will be kept private. 
The performance on these datasets  determined the winners of the challenge. 
\end{itemize}



\subsection{Schedule}
The competition adhered to  the following schedule.
\begin{itemize}
\item \textbf{30th July, 2018.} Beginning of the competition, release of development data. Participants started submitting code and obtaining immediate feedback in the leaderboard. 
\item \textbf{30th October, 2018:} End of development (Feedback) phase.  
\item \textbf{6th November, 2018:} End of the competition. Code from phase 1 was migrated automatically to phase 2. 
Code was verified by organizers. 
\item \textbf{13th November, 2018:} Deadline for submitting the fact sheets. 
\item \textbf{20th November, 2018.} Release of results, announcements of winners. 


\item \textbf{December 7th, 2018.} NeurIPS 2018 Autonomous Lifelong Machine Learning with Drift competition session, dissemination of results, award ceremony.
\end{itemize}

\subsection{Prizes}
\label{sec:prizes}
We offered cash prizes and travel grants to the top ranked participants of the challenge. Prizes donated by 4Paradigm were distributed as follows:
\begin{itemize}
    \item First Place Prize: \$10,000USD
	\item Second Place Prize: \$3,000USD
	\item Third Place Prize: \$2,000USD
\end{itemize}
Additionally, travel grants sponsored by ChaLearn of \$1,000USD each were offered to participants for attending the competition session at the NeurIPS conference. 

Since two competition environments were provided to participants, each with its own website (see Section~\ref{sec:platform}) at the end of the competition we had two leader boards. Both leader boards were merged and the top ranked participants were considered to be eligible for prizes. In order to be able to receive the prize, participants had to release their solution under any public license. Also, they had to describe in detail their solution in a fact sheet and present their work in the competition session at NeurIPS2018. 

\section{Overview of results}
\label{sec:results}
The feedback phase lasted about 3 months and attracted more than 300 \emph{unique} participants, see Table~\ref{tab:stats} for a summary of participation statistics.  We distinguish registered from active participants, as the former only registered to the competition, whereas the latter actively participated in the challenge making code submissions to either of the available competition settings (i.e., Python-2 or Python-3). Almost 2,000 submissions were received during the feedback phase, representing about 4,000 hours of computing time (assuming submissions adhere to the budget). This does not include the time spend in the final evaluation phase. 
\begin{table}[htb]
	\centering
	\caption{Participation statistics.}
	\label{tab:stats}\footnotesize{
	\begin{tabular}{|c|c|c|c|}
		\hline
		Duration  &Registered participants & Active participants & Submissions Feedback phase \\ \hline
		90 Days& 334       & $\approx$100 & $\approx$ 1,800\\\hline
	\end{tabular}}
\end{table}

At the end of the feedback phase the last successful submission from each participant was migrated into the  final phase. In this phase code was evaluated in  fresh new datasets that were hidden for all participants. Performance in these private datasets was used to determine the winners. Although participants did not know anything from these datasets, these were very similar to those used in the feedback phase. 

A leader board per each setting, Python-2 and Python-3,  was generated by considering submissions that succeeded in processing all of the datasets from the final phase. From the Python-2 and Python-3 leader boards a single and final leader board was generated. These results were used to rank participants and distribute prizes.  Submissions coming form participants that regularly made submissions to both, Python-2 and Python-3, sites were removed from the final leader board.  

The official merged ranking is shown in Table~\ref{tab:ranking}. We show the rank of each valid submission in the five private datasets, the average ranking which was the leading evaluation measure and the duration in seconds from each of the submissions. Please note that we are not revealing the actual performance of methods in the datasets because these are being used in the AutoML rematch  challenge. 
\begin{table}[htb]
	\centering
	\caption{Final leader board of the challenge.}
	\label{tab:ranking}\tiny{
	\begin{tabular}{p{0.7cm}|p{0.7cm}|p{1.5cm}|c|c|c|c|c|c|c}
		\hline
		\textbf{Ranking}	&\textbf{Bundle}	&	\textbf{Team Name}	&	\textbf{Avg. rank}	&	\textbf{Set 1}	&	\textbf{Set 2}	&	\textbf{Set 3}	&	\textbf{Set 4}	&	\textbf{Set 5}	&	\textbf{Duration}	\\\hline
1	&	Py3	&	autodidact.ai	&	2.2	&	2	&	4	&	1	&	2	&	2	&	5882.13	\\
2	&	Py3	&	Meta\_Learners	&	2.4	&	3	&	1	&	2	&	1	&	5	&	8700.47	\\
3	&	Py3	&	GrandMasters	&	4.2	&	4	&	6	&	4	&	3	&	4	&	7912.14	\\
4	&	Py3	&	Ml-Intelligence	&	4.2	&	1	&	3	&	6	&	10	&	1	&	9426.68	\\
5	&	Py3	&	linc326	&	4.6	&	6	&	5	&	5	&	4	&	3	&	8843.15	\\
6	&	Py3	&	rcarson	&	6.4	&	9	&	2	&	7	&	6	&	8	&	5471.59	\\
7	&	Py3	&	jimliu	&	7.8	&	5	&	8	&	15	&	5	&	6	&	5581.74	\\
8	&	Py3	&	PGijsbers	&	8.4	&	7	&	7	&	14	&	7	&	7	&	10427.18	\\
9	&	Py3	&	gxr\_6666	&	11.4	&	13	&	14	&	10	&	8	&	12	&	6674.08	\\
10	&	Py2	&	pipi\_	&	12	&	10	&	10	&	8	&	18	&	14	&	8334.86	\\
11	&	Py3	&	Jie\_NJU	&	12	&	11	&	23	&	3	&	13	&	10	&	8282.08	\\
12	&	Py2	&	nomo	&	15	&	14	&	15	&	18	&	12	&	16	&	4165.99	\\
13	&	Py3	&	mlg.postech	&	16.2	&	12	&	26	&	9	&	23	&	11	&	7357.6	\\
14	&	Py2	&	Cheng\_Zi	&	18.4	&	19	&	17	&	19	&	22	&	15	&	4532.53	\\
15	&	Py2	&	hugo.jair	&	19.8	&	20	&	20	&	21	&	21	&	17	&	3554	\\
16	&	Py2	&	eric2	&	22	&	16	&	19	&	23	&	17	&	35	&	3459.95	\\
17	&	Py3	&	ckirby	&	23	&	58	&	9	&	30	&	9	&	9	&	5105.47	\\
18	&	Py3	&	aad\_freiburg	&	23.6	&	8	&	16	&	20	&	27	&	47	&	10411.68	\\
19	&	Py3	&	MichaelIbrahim	&	24	&	34	&	11	&	11	&	14	&	50	&	2875.06	\\
20	&	Py3	&	FHM	&	25	&	35	&	12	&	12	&	15	&	51	&	2823.64	\\
21	&	Py3	&	GaoGege	&	25.4	&	36	&	13	&	13	&	16	&	49	&	2881.66	\\
22	&	Py2	&	Xiangning	&	26	&	17	&	21	&	16	&	30	&	46	&	3925.01	\\
23	&	Py3	&	Vamshidhar	&	26.6	&	48	&	27	&	34	&	11	&	13	&	5315.57	\\
24	&	Py2	&	kong	&	27.4	&	15	&	37	&	17	&	20	&	48	&	3747.17	\\
25	&	Py2	&	naiven	&	28.2	&	22	&	33	&	28	&	38	&	20	&	4822.63	\\
26	&	Py2	&	ninad	&	28.4	&	18	&	24	&	24	&	35	&	41	&	6096.44	\\
27	&	Py2	&	leogautheron	&	29.4	&	21	&	22	&	22	&	39	&	43	&	3662.37	\\
28	&	Py2	&	gbramble	&	30	&	24	&	31	&	31	&	36	&	28	&	3654.6	\\
29	&	Py2	&	iamrobot	&	31	&	25	&	32	&	32	&	37	&	29	&	3782.51	\\
30	&	Py2	&	johnnytorres83	&	34.6	&	23	&	28	&	33	&	50	&	39	&	3451.06	\\
31	&	Py2	&	amitmeher	&	35.6	&	42	&	35	&	37	&	46	&	18	&	3453.64	\\
32	&	Py2	&	philipjhj	&	35.4	&	30	&	48	&	44	&	24	&	31	&	3297.31	\\
33	&	Py3	&	yanzhen0923	&	36.8	&	43	&	18	&	47	&	19	&	57	&	3944.87	\\
34	&	Py2	&	fanqie	&	37.2	&	28	&	34	&	27	&	43	&	54	&	5350.05	\\
35	&	Py2	&	mvslee	&	38.4	&	38	&	25	&	25	&	51	&	53	&	3799.02	\\
36	&	Py2	&	ohmygirl	&	38.2	&	29	&	50	&	56	&	32	&	24	&	3279.1	\\
37	&	Py2	&	xiayunsun	&	40	&	26	&	59	&	29	&	60	&	26	&	3336.72	\\
38	&	Py2	&	ObserverL	&	40	&	51	&	54	&	39	&	33	&	23	&	3261.85	\\
39	&	Py2	&	trevin.gandhi	&	40.4	&	41	&	46	&	60	&	25	&	30	&	3180.56	\\
40	&	Py2	&	kim.putin	&	40.6	&	46	&	43	&	59	&	28	&	27	&	3206.82	\\
41	&	Py3	&	lhg1992	&	41	&	49	&	30	&	36	&	31	&	59	&	2702.37	\\
42	&	Py2	&	water\_water	&	41.8	&	61	&	39	&	42	&	42	&	25	&	4501.96	\\
43	&	Py2	&	derplearning	&	42.6	&	33	&	40	&	26	&	59	&	55	&	4797.65	\\
44	&	Py3	&	TAU	&	42.4	&	50	&	38	&	35	&	29	&	60	&	2493.08	\\
45	&	Py2	&	cindy\_y	&	43.2	&	27	&	60	&	49	&	61	&	19	&	3247.28	\\
46	&	Py2	&	Zhengying	&	43.2	&	37	&	61	&	52	&	34	&	32	&	3190.33	\\
47	&	Py2	&	OhYeah	&	43.6	&	57	&	52	&	61	&	26	&	22	&	3317.84	\\
48	&	Py2	&	jiaorunnju	&	44	&	47	&	49	&	41	&	47	&	36	&	3247.42	\\
49	&	Py2	&	hcilab	&	44.4	&	56	&	42	&	38	&	53	&	33	&	3235.11	\\
50	&	Py2	&	prabhant	&	45.2	&	31	&	57	&	45	&	41	&	52	&	3216.81	\\
51	&	Py2	&	ailurus	&	45.4	&	39	&	47	&	55	&	52	&	34	&	3168.49	\\
52	&	Py2	&	yush	&	45.8	&	54	&	55	&	54	&	45	&	21	&	3363.55	\\
53	&	Py2	&	cyxlily	&	46.2	&	32	&	53	&	57	&	49	&	40	&	4413.06	\\
54	&	Py2	&	Malik	&	46.4	&	40	&	56	&	43	&	56	&	37	&	3334.81	\\
55	&	Py2	&	bestever	&	46.4	&	53	&	51	&	50	&	40	&	38	&	3225.51	\\
56	&	Py2	&	hx173149	&	47	&	44	&	36	&	51	&	48	&	56	&	3879.95	\\
57	&	Py2	&	utpalsikdar	&	47.2	&	52	&	58	&	40	&	44	&	42	&	3192.67	\\
58	&	Py2	&	sherryxue1991	&	47.4	&	45	&	29	&	48	&	57	&	58	&	7980.14	\\
59	&	Py2	&	Xiang\_Liu	&	51.2	&	60	&	44	&	53	&	55	&	44	&	3195.86	\\
60	&	Py2	&	ostapeno	&	51.6	&	55	&	41	&	46	&	54	&	62	&	3261.26	\\
61	&	Py2	&	kongyanye	&	53	&	59	&	45	&	58	&	58	&	45	&	3239.34	\\
62	&	Py2	&	JimmyChang	&	61.8	&	62	&	62	&	62	&	62	&	61	&	5263.46	\\
63	&	Py2	&	zhqiu	&	63	&	63	&	63	&	63	&	63	&	63	&	3151.24	\\\hline
	\end{tabular}}
\end{table}

Initially a total of 103 submissions were considered to build the final leader board (68 coming from the Python-2 competition setting and 35 from the Python-3 one). After filtering out invalid submissions and discarding submissions that failed to process the five private datasets we ended up with the 63 valid submissions listed in Table~\ref{tab:ranking}.   
From this table it can be seen that the  top-9 teams in this ranking used the Python-3 competition setting, showing the impact that incorporating it  had into the final results. Although the best positioned  Python-2   entry outperformed several Python-3 submissions. 

As previously mentioned  there were several top ranked submissions in the feedback phase that failed to process the datasets from the final phase. Different errors arose at this stage, mostly related with exceeding the available resources. In all cases we reproduced the errors offline, some of these submissions successfully finished but using much more resources than those available to participants. In Table~\ref{tab:rakingfeedback} we show for reference the top 10 ranked submissions in the feedback phase.  
\begin{table}[htb]
	\centering
	\caption{Top 10 ranked participants in the feedback phase.}
	\label{tab:rakingfeedback}\tiny{
	\begin{tabular}{p{0.7cm}|p{1.5cm}|c|c|c|c|c|c|c}
		\hline
\textbf{Rank}	&	\textbf{Participants}	&	\textbf{Avg. rank}	&	\textbf{A}	&	\textbf{B}	&	\textbf{C}	&	\textbf{D}	&	\textbf{E 5}	&	\textbf{Duration}	\\\hline
1	&	deepsmart	&	1.2	&	0.5614 (1)	&	0.3489 (2)	&	0.6216 (1)	&	0.6027 (1)	&	0.8112 (1)	&	6167.27	\\
2	&	HANLAB	&	3	&	0.5344 (5)	&	0.3372 (4)	&	0.5815 (2)	&	0.5676 (2)	&	0.7848 (2)	&	7289.04	\\
3	&	Fong	&	4.2	&	0.5370 (4)	&	0.3356 (5)	&	0.5806 (3)	&	0.5561 (5)	&	0.7795 (4)	&	6555.01	\\
4	&	Ml-Intelligence	&	4.4	&	0.5456 (2)	&	0.3539 (1)	&	0.4874 (10)	&	0.5443 (6)	&	0.7829 (3)	&	7313.47	\\
5	&	QQSong	&	4.6	&	0.5372 (3)	&	0.3306 (6)	&	0.5545 (6)	&	0.5667 (3)	&	0.7712 (5)	&	6172.42	\\
6	&	tnguyen	&	4.6	&	0.5294 (6)	&	0.3442 (3)	&	0.5683 (4)	&	0.5643 (4)	&	0.7532 (6)	&	6936.53	\\
7	&	autodidact.ai	&	7.2	&	0.5171 (7)	&	0.3088 (9)	&	0.5645 (5)	&	0.4779 (8)	&	0.7273 (7)	&	4551.58	\\
8	&	Meta\_Learners	&	8.2	&	0.4924 (8)	&	0.3104 (8)	&	0.5463 (7)	&	0.4780 (7)	&	0.6943 (11)	&	6365.39	\\
9	&	linc326	&	8.8	&	0.4641 (9)	&	0.3239 (7)	&	0.4768 (11)	&	0.4744 (9)	&	0.7070 (8)	&	7101.69	\\
10	&	GrandMasters	&	10	&	0.4632 (11)	&	0.2878 (10)	&	0.5033 (8)	&	0.4578 (11)	&	0.7048 (10)	&	5981.12	\\\hline
\end{tabular}}
\end{table}

Two observations can be made (1) some participants seem to have overfitted the leader board (i.e., compare the relative performance  of \emph{Ml-intelligence} and \emph{linc326} with \emph{autodidact.ai} in Tables~\ref{tab:ranking} and Table~\ref{tab:rakingfeedback}), and (2) submissions from half of the top ranked participants in the feedback phase failed to process the private data. Thus, for a large fraction of the participant, this task of automated machine learning with drift was very hard.

The top 3 ranked submissions in Table~\ref{tab:ranking} were eligible for prizes (see Section~\ref{sec:prizes}). Please note that as there was a tie in average ranking of entries by \emph{Ml-Intelligence} and \emph{linc326} teams, the duration was used as tie breaking criterion.  Top ranked teams were asked to provide a fact sheet describing their solution and releasing their code under any public license. After code verification and ensuring that eligible teams complied with the rules, prizes were awarded to the top 3 ranked teams as follows: 
\begin{itemize}
    \item \textbf{1st place.  Autodidact.ai}. \emph{Jobin Wilson, Amit Kumar Meher, Bivin Vinodkumar Bindu, Manoj Sharma, Vishakha Pareek.} Flytxt, Indian Institute of Technology Delhi, CSIR-CEERI
    \item \textbf{2nd place. Meta\_Learners.}  \emph{Zheng Xiong, Jiyan Jiang, Wenpeng Zhang} Tsinghua University, China
    \item \textbf{3rd place. GrandMasters. }  \emph{Jiangeng Chang, Yakun Zhao, Honggang Liu, Jinlong Chai}. BeiJing University of Post and Telecom WCSN Lab, BeiJing University of Post and Telecom AI \& HPC Department. Inspur Electronic Central South University, China
\end{itemize}

A brief description of solutions proposed by these teams is provided in Table~\ref{tab:solutions}. It can be seen that, similarly to the provided baseline, participants relied on a boosting tree ensemble. Thus we may question whether the baseline method may have biased the challenge. Another common technique used by the top ranking participant was an efficient coding of categorical and multi-valued categorical features. Subsampling was implemented by the three teams, with different subsampling approaches being considered. Finally, two out of the three top ranked teams used hyperparameter optimization. 
\begin{table}[htb]
\centering
\caption{Overview of proposed solutions by top ranked participants. GBDT: Gradient boosting decision tree; MVC: Multi-valued categorical; SMBO: sequential model-based global optimization.}
\label{tab:solutions}
\tiny{
\begin{tabular}{| p{0.5cm} |  p{1.6 cm} |  p{2cm} | p{2cm} | p{2cm} | p{2cm} | }
\hline
\textbf{Pos.} &  \textbf{Predictive Model} & \textbf{Feature processing} & \textbf{Concept drift mechanism} & \textbf{Subsampling}& \textbf{Hyperparameter optimization} \\ \hline
1& GBDT Ensemble (ligthlight GBT) & Statistics derived from data, count based encoding for Cat. and MVC feats. & Storage of historical data& Subsampling and storage of historical data&SMBO\\\hline
2&GBDT Ensemble & Count encoding and target conditional encoding for Cat and MVC feats. Co-encoding: Training and test instances encoded in a single step. & Incremental learning  &Selective subsampling, recent samples preferred&Optimization of hyperparameters restricted on budget\\\hline
3& GBDT Ensemble  & Double encoding for categorical and MVC feats.: ordinal and count based& Incremental learning with adaptive learning rate & Sliding window subsampling: last two batches of data are retained & Learning rate was modified with each batch\\\hline
\end{tabular}}
 \end{table}

Regarding the rest of participants, in the following we summarize statistics collected from those participants that filled and sent a fact sheet describing their solution. This summary is based on the fact sheets from 21 teams (85\% of these coming form participants working in the Python-3 bundle, and the rest from the Python-2 one). 

\begin{figure}[htb]
    \centering
    \includegraphics[width=0.48\textwidth]{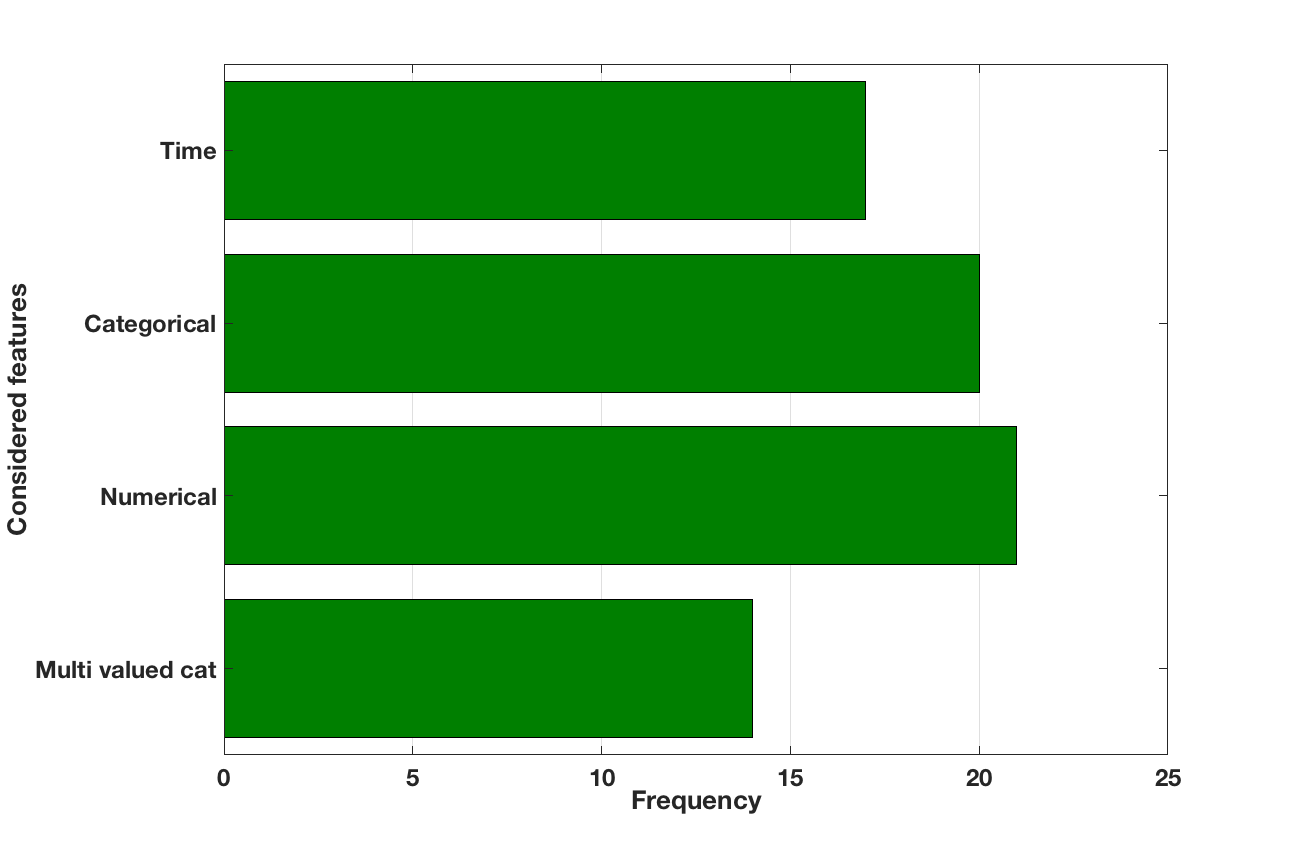}
    \includegraphics[width=0.48\textwidth]{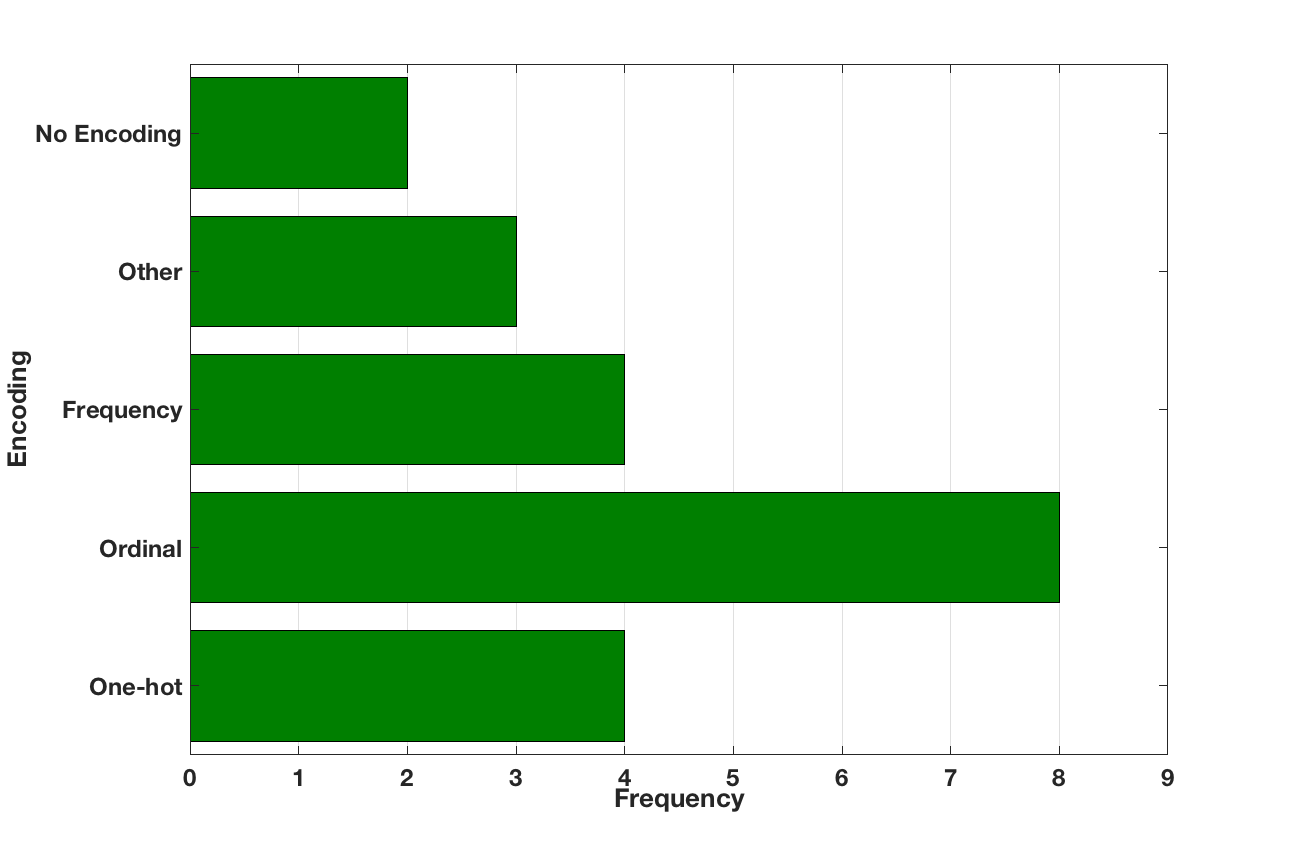}
    \vspace{-0.3cm}
    \caption{Feature usage statistics.}
    \label{fig:featusage}
\end{figure}

Figure~\ref{fig:featusage} shows an histogram on the usage of the different types of available features (left) and the encoding mechanism adopted for categorical and multi valued categorical variables (right). Everyone used numerical features and almost everyone (20 out of 21) considered categorical variables. Multi valued attributes were the less used ones, this could be due to the difficulty associated to encode this sort of attributes, together with the computational load involved in  loading them.  Most participants used ordinal encoding when using categorical and multi valued categorical features, this was the encoding method provided with the baseline. The reason for this could be the fact that this is among the most efficient encoding methods. 

Regarding predictor, it is quite interesting that all participants that sent the fact sheet have reported to use boosting decision tree based ensemble as predictor. Few participants used this model in combination with another one (either a linear model or a Gaussian process). 
This trend is in agreement with other challenges\footnote{\url{https://www.kdnuggets.com/2017/10/xgboost-top-machine-learning-method-kaggle-explained.html}}~\cite{automlchallenges},   and in general with supervised learning trends.  It is also interesting that even when this was an AutoML challenge most participants did not perform an optimization of hyperparameters, this could be due to the limited resources and the size of the datasets.

Regarding the strategies adopted by participants to deal with concept drift and the lifelong setting we can find:
\begin{itemize}
    \item Incrementally adding models to an ensemble 
    \item Hyperparameter optimization in different batches
    \item Keeping and subsampling data from different batches
    \item Dynamic feature encoding (across batch) 
    \item Model training in the last previous batch
    \item Curriculum learning
\end{itemize}

The previous summary gives an idea on the variety of techniques for dealing with drift and the sequential evaluation being implemented by participants. Although we are still far away from solving the approached problem, it is clear that the challenge has boosted research in the topic and will surely will motivate novel methodologies that can cope with the difficulties associated with the challenge. 
\section{Final remarks}
\label{sec:discussion}
We have described the \emph{Autonomous Lifelong Machine Learning with Drift} challenge that was collocated with the NeurIPS2018 conference. This competition challenged participants to develop autonomous supervised learning machines capable of dealing with concept drift in a relaxed lifelong learning and evaluation setting. In addition the competition considered huge datasets and limited resources, increasing the difficulty of the task. We adopted an evaluation protocol that allowed the automated evaluation of  solutions.  

Overall, the challenge attracted a vast number of participants, and motivated the adaptation of existing methods and development of new methodologies for dealing with concept drift in a lifelong setting. Interestingly, most participants used a similar predictive model, which was updated in different ways. An aspect of decisive relevance was that of feature characterization and efficiency of solutions. The organized challenge has motivated rematches considering the same setting. In the following we summarize the main conclusions and findings derived from the competition:
\begin{itemize}
    \item The \textbf{competition environment makes a difference} in the overall results of the challenge. The suggestion from participants to upgrade the competition bundle to account for a more recent environment paid off. It was clear from results that better performance as obtained by participants that used the Python-3 setting. 
    \item \textbf{Resource and time handling is critical for AutoML and code submission competitions}. Several top ranked participants failed to process the datasets from the final phase within the associated budget. In this regard, solutions offering a good tradeoff between model complexity and performance would be preferred for this sort of challenges.
    \item \textbf{Large datasets prevented participants from focusing on other aspects of AutoML}. The size of the datasets and the limited resources made participants to struggle with defining efficient data-loading feature-coding mechanisms. This could prevent several participants to use standard hyperparameter optimization / AutoML methods, e.g., AutoSKLearn~\cite{autosklearn,madrid:hal-01966962}; although this could also be due to the robustness of the considered predictors together with the size of data sets. 
    \item Most solutions converged to \textbf{similar pipelines} that comprised: feature characterization, subsampling and data storage, and incremental update of a predictor (a boosting ensemble of trees). However a variety of mechanisms were adopted to deal with the drift and the lifelong evaluation setting. 
    \item Adequate encoding and representation of categorical and multi value categorical variables is important. 
    \item Storing historical data paids off. 
\end{itemize}

While the challenge succeeded in attracting the attention from the community, this  was just the start. The proposed challenge setting is one of practical relevance, likewise,  the size of the datasets together with the involved features are representative of the type of problems one can find in practice. Much remains to be done, in particular, \textbf{AutoML methods for automatically encoding heterogeneous features} and extracting relevant discriminative information would have a great impact. Another promising venue for research is that of studying AutoML solutions able to work incrementally, i.e., \textbf{online AutoML}. In broader terms, another step towards full automation would involve designing methods that are able to work from raw data directly. In this direction, a new series of challenges targeting deep learning (\textbf{AutoDL}) will boost research on AutoML of broader impact, see~\cite{liu:hal-01906226}.

\section*{Acknowledgments}
The authors are grateful with the challenge sponsors 4Paradigm, Microsoft Research and ChaLearn, as well as  the NeurIPS2018 competition chairs. 
\bibliographystyle{plain}

\end{document}